\documentclass[10pt,twocolumn,letterpaper]{article}

\usepackage[pagenumbers]{cvpr} 

\usepackage{graphicx}
\usepackage{amsmath}
\usepackage{amssymb}
\usepackage{booktabs}
\usepackage{array}
\usepackage{tabularx}
\usepackage{pifont}
\usepackage{color, colortbl}

\usepackage{booktabs, multirow, makecell} 
\usepackage{soul}
\usepackage[table]{xcolor} 

\usepackage{amssymb}
\usepackage{pifont}

\definecolor{Gray}{gray}{0.9}

\usepackage[pagebackref,breaklinks,colorlinks]{hyperref}

\usepackage[capitalize]{cleveref}
\crefname{section}{Sec.}{Secs.}
\Crefname{section}{Section}{Sections}
\crefname{table}{Tab.}{Tabs.}
\Crefname{table}{Table}{Tables}

\def\cvprPaperID{xxxx} 
\def\confName{CVPR}
\def\confYear{2022}

\title{JRDB-Traj: A Dataset and Benchmark for Trajectory Forecasting in Crowds \\ \small Technical report
}

\author{Saeed Saadatnejad$^{1}$, Yang Gao$^{1}$, Hamid Rezatofighi$^{2}$ and Alexandre Alahi$^{1}$\\
$^{1}$EPFL, $^{2}$Monash University \\ 
{\tt\small \{saeed.saadatnejad\}@epfl.ch, \{hamid.rezatofighi\}@monash.edu}
}
\begin{document}

\makeatother
\twocolumn[{
\maketitle
\begin{center}
    \captionsetup{type=figure}
\includegraphics[width=\textwidth]{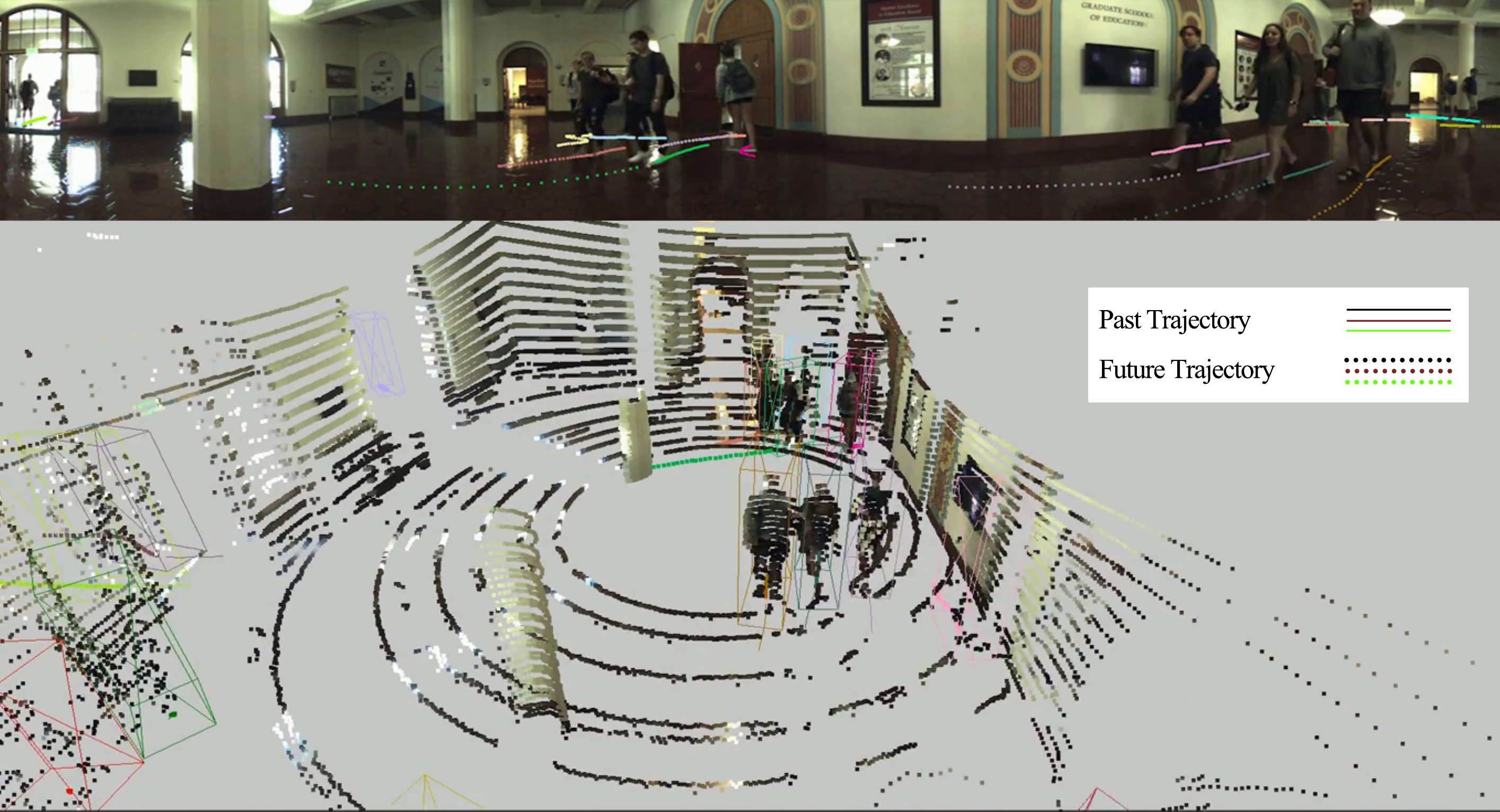}
    \captionof{figure}{Trajectory Forecasting in JRDB-Traj dataset. The top figure displays the RGB image from the robot's perspective, while the bottom figure shows the corresponding 3D point cloud. Solid lines indicate the observed past trajectories and dots represent the ground-truth future trajectories.}
    \label{fig:pull_figure}
\end{center}
}]

\begin{abstract}
Predicting future trajectories is critical in autonomous navigation, especially in preventing accidents involving humans, where a predictive agent's ability to anticipate in advance is of utmost importance. Trajectory forecasting models, employed in fields such as robotics, autonomous vehicles, and navigation, face challenges in real-world scenarios, often due to the isolation of model components. To address this, we introduce a novel dataset for end-to-end trajectory forecasting, facilitating the evaluation of models in scenarios involving less-than-ideal preceding modules such as tracking. This dataset, an extension of the JRDB dataset, provides comprehensive data, including the locations of all agents, scene images, and point clouds, all from the robot's perspective. The objective is to predict the future positions of agents relative to the robot using raw sensory input data.
It bridges the gap between isolated models and practical applications, promoting a deeper understanding of navigation dynamics. Additionally, we introduce a novel metric for assessing trajectory forecasting models in real-world scenarios where ground-truth identities are inaccessible, addressing issues related to undetected or over-detected agents. Researchers are encouraged to use our benchmark for model evaluation and benchmarking.
\end{abstract}

\section{Introduction}

The ability to predict future events is widely regarded as a fundamental aspect of intelligence~\cite{bubic2010prediction}. This predictive capability assumes paramount significance in the context of autonomous navigation, where precise predictions play a pivotal role in preventing accidents involving humans. A predictive agent possesses the foresight to anticipate the agents' actions few seconds in advance, enabling it to make informed decisions about when to stop or proceed safely.
Trajectory forecasting models are aimed at predicting the future positions of agents based on a sequence of past observed locations. These models have been used in socially-aware robotics~\cite{chen2019crowd}, autonomous vehicles~\cite{saadatnejad2022sattack}, and navigation~\cite{luo2018porca}.

With the successful development of deep learning, autonomous navigation algorithms have undergone a significant transformation, involving a complex series of interconnected tasks such as object detection, tracking, and trajectory forecasting. In the realm of industry solutions, it is common to deploy standalone models for each of these tasks, often evaluating and comparing them independently without considering their interdependencies. In other words, it assumes that previous modules function optimally. In practice, early modules may exhibit imperfections that can result in less-than-ideal trajectory forecasting.

We present a novel dataset designed for end-to-end trajectory forecasting in order to study the performance given non-perfect previous modules such as tracking. This is an extension of JRDB~\cite{martin2021jrdb} dataset viewed from the perspective of a robot navigating within a dynamic environment. The task is to predict the future positions of all agents within the scene relative to the robot, leveraging raw sensory input data, point clouds and images. By focusing on end-to-end trajectory forecasting, our goal is to bridge the gap between isolated models and practical applications, fostering a deeper understanding of real-world navigation dynamics.

Notably, assessing the future trajectory prediction performance for multiple agents poses a challenge. Common metrics such as Average Displacement Error (ADE) and Final Displacement Error (FDE) cannot be employed because complete and accurate observed trajectories are unavailable. In essence, we lack the associated identities (IDs) required to calculate the displacement error accurately. Moreover, the inclusion of detection and tracking models can lead to instances where agents are either not detected or are over-detected, subsequently affecting the input data provided to the forecasting model.
To address these issues, we introduce a novel, comprehensive metric for evaluating trajectory forecasting models in a two-step process involving matching and measuring displacement.
We made our benchmark publicly accessible, inviting researchers to submit their trajectory prediction models for evaluation and benchmarking against this new metric.

\section{The JRDB-Traj Dataset}
JRDB~\cite{martin2021jrdb}, JRDB-Act~\cite{ehsanpour2022jrdb} and JRDB-Pose~\cite{vendrow2023jrdb} previously introduced annotations including 2D and 3D bounding boxes with tracking IDs, action labels, social groups and body pose.
We leverage the 3D bounding box annotations and make the trajectories using the center of the bounding box of the person on the ground.

\subsection{Splits}
We follow the official splits of JRDB~\cite{martin2021jrdb} to create training, validation, and testing splits from the 54 captured sequences, with each split containing an equal proportion of indoor and outdoor scenes as well as scenes captured using a stationary or moving robot. All frames from a scene appear strictly in one split.
The videos and point clouds for the last five seconds of the test are hidden.

\subsection{Evaluation Metrics}
Assessing trajectory forecasting performance in the absence of ground-truth IDs necessitates the establishment of associations between predicted and ground-truth trajectories in future frames, followed by their distance measurement—a standard approach in detection and tracking evaluations. Therefore, we report two prevalent detection and tracking metrics in these future frames:
\begin{enumerate}
\item IDF1~\cite{ristani2016idf1}: This is the ratio of correctly identified detections over the average number of ground-truth and computed detections.
\item OSPA-2~\cite{rezatofighi2020trustworthy}: Optimal Sub-Pattern Matching (OSPA)~\cite{schuhmacher2008consistent} is a multi-object performance evaluation metric which includes the concept of miss-distance in tracking. OSPA-2 has been further adapted to detection and tracking tasks. It is a set-based metric that can directly capture a distance between two sets of trajectories without a thresholding parameter.
\end{enumerate} 

Furthermore, we propose End-to-end Forecasting Error (EFE) for assessing trajectory forecasting in real-world scenarios. In short, EFE determines the associations between predicted and ground-truth trajectories, measures their distances, and accounts for any mismatches in the number of trajectories. 
Importantly, EFE refrains from penalizing early terminations in predicted trajectories. In practical terms, if a model predicts a trajectory for an agent that extends beyond the scene boundaries, it does not contribute to error. A comprehensive explanation follows.

Let $\mathbf{X} = \{X^{\mathcal{D}_1}_1, X^{\mathcal{D}_2}_2, \dots X^{\mathcal{D}_m}_m\}$ and $\mathbf{Y} = \{Y^{\mathcal{D}_1}_1, Y^{\mathcal{D}_2}_2, \dots Y^{\mathcal{D}_n}_n\}$ be the sets of trajectories for prediction and the ground-truth, respectively. Note $\mathcal{D}_i$ represents the time indices which track $i$ exists (having a state-value). Then, we calculate the time average distance of every pair of tracks $X^{\mathcal{D}_i}_i$ and $Y^{\mathcal{D}_j}_j$:
\begin{equation}
\label{eq:time-av}
\underline{\widetilde{d}}(X^{\mathcal{D}_i}_i, Y^{\mathcal{D}_j}_j)
=\sum_{t \in \mathcal{D}_i \cup \mathcal{D}_j} \frac{d_{O}\left(\{X_i^{t}\},
\{Y_j^{t}\}\right)}{|\mathcal{D}_{i} \cup \mathcal{D}_{j}|},
\end{equation}
where $t \in \mathcal{D}_{i} \cup \mathcal{D}_{j}$ is the time-step when either or both track presents. Note that $\{X_i^{t}\}$ and $\{Y_j^{t}\}$ are singleton sets, \ie $\{X_i^{t}\} = \emptyset$ or $\{X_i^{t}\} = x_i^t\in \mathbb{R}^{2}$ and $\{Y_j^{t}\} = \emptyset$ or $\{Y_j^{t}\} = y_j^t\in \mathbb{R}^{2}$. Therefore, $d_{O}\left(\{X_i^{t}\} , \{Y_j^{t}\}\right)$ can be simplified into the following distance function, :
\begin{align}
d_{O}(\{X_i^{t}\},&\{Y_j^{t}\})= \nonumber\\
&\left\{\begin{array}{lll}
d_c(x^t_i, y^t_j) &\text{if } |\{X_i^{t}\}|\wedge|\{Y_j^{t}\}|=1, \\
c &\text{if } |\{X_i^{t}\}|=0 \;\&\; |\{Y_j^{t}\}| \;!=0,\\
0 &\text{Otherwise,}
\end{array}\right.
\end{align}
where $d_{c}(x_i, y_i) := min(c, d(x_i, y_i))$ indicates the euclidean displacement error with the cutoff distance $c$.

Finally, we obtain the distance, EFE, between two sets of trajectory tracks, \ie $\mathbf{X}$ and $\mathbf{Y}$ by:
\begin{align}
\label{eq:efe1}
 EFE&(\mathbf{X},\mathbf{Y})=\nonumber\\ \frac{1}{n}&\left(\min_{\pi \in \Pi_{n}} \sum_{i=1}^{m}\underline{\widetilde{d}}(X^{\mathcal{D}_i}_i, Y^{\mathcal{D}_{\pi_i}}_{\pi_i})+c*(n-m)\right),
\end{align}
if $n \geq m$. $\Pi_n$ is the set of all permutations of $\{1, 2, \dots, n\}$. 
Note that $\underline{\widetilde{d}}$ reflects the localization errors of trajectories, whereas $c*(n-m)$ reflects the cardinality error (false and missed trajectories) and we put $c=5$ meters as the threshold penalty.
If $m > n$:
\begin{align}
\label{eq:efe2}
 EFE&(\mathbf{X},\mathbf{Y})=\nonumber\\ \frac{1}{m}&\left(\min_{\pi \in \Pi_{m}} \sum_{i=1}^{n}\underline{\widetilde{d}}(X^{\mathcal{D}_{\pi_i}}_{\pi_i}, Y^{\mathcal{D}_i}_i)+c*(m-n)\right).
\end{align}
We further define $EFE(X, Y) = c$ if either $X$ or $Y$ is empty, and $EFE(\emptyset, \emptyset) = 0$. 
The code of the metrics and related details can be accessed in \href{https://github.com/JRDB-dataset/jrdb_toolkit}{JRDB Toolkit}.

\subsection{Benchmark}
In \Cref{tab:results}, we have evaluated the performance of the well-known Social-LSTM baseline~\cite{alahi2016social} in addition to the simple Zero-Velocity baseline, which repeats the last observed location of each agent as its future predicted locations.
Note that we forecast future trajectories by leveraging observed past trajectory estimates derived from detection and tracking algorithms. Here, we utilized the estimated detections by a pre-trained PiFeNet~\cite{le2022pifenet} and subsequently employed the Simpletrack~\cite{pang2022simpletrack} tracking algorithm to provide observed past trajectory estimates as inputs to all the aforementioned models.
We employed the Social-LSTM code provided by TrajNet++~\cite{kothari2021human}.
Nevertheless, it is essential to note that we predicted an agent's future trajectory when we had access to the last two observed data points within their trajectory.
Our code is available online: \href{https://github.com/vita-epfl/JRDB-Traj}{https://github.com/vita-epfl/JRDB-Traj}.

\begin{table}[!t]
    \centering
    \begin{tabular}{|lccc|}
    \hline
        Model &  EFE $\downarrow$ & OSPA-2 $\downarrow$ & IDF1 $\uparrow$ \\
        \hline
        Zero-Velocity & 2.981 & 3.082 & 49.431\\
        Social-LSTM~\cite{alahi2016social} & 2.646 & 2.76 & 54.673\\
        \hline
    \end{tabular}
    \caption{Quantitative evaluations of trajectory forecasting models on JRDB-Traj dataset.}
    \label{tab:results}
\end{table}

\section{Conclusion}
In this paper, we introduced JRDB-Traj, a new dataset and benchmark for trajectory forecasting from raw sensory inputs.
We have also introduced EFE, a new metric for trajectory forecasting in crowds where ground-truth identities are inaccessible.
We anticipate that this dataset will foster further research in this domain, bringing us closer to realizing a fully functional autonomous navigation system suitable for practical applications.

\section*{ACKNOWLEDGMENT}
The authors thank Simindokht Jahangard and Duy Tho Le for their valuable input.



{\small
\bibliographystyle{ieee_fullname}
\bibliography{ref}

\begin{thebibliography}{10}\itemsep=-1pt

\bibitem{alahi2016social}
Alexandre Alahi, Kratarth Goel, Vignesh Ramanathan, Alexandre Robicquet, Li
  Fei-Fei, and Silvio Savarese.
\newblock Social lstm: Human trajectory prediction in crowded spaces.
\newblock In {\em Proceedings of the IEEE/CVF conference on computer vision and
  pattern recognition (CVPR)}, pages 961--971, 2016.

\bibitem{bubic2010prediction}
Andreja Bubic, D.~Yves Von~Cramon, and Ricarda Schubotz.
\newblock Prediction, cognition and the brain.
\newblock {\em Frontiers in Human Neuroscience}, 4:25, 2010.

\bibitem{chen2019crowd}
Changan Chen, Yuejiang Liu, Sven Kreiss, and Alexandre Alahi.
\newblock Crowd-robot interaction: Crowd-aware robot navigation with
  attention-based deep reinforcement learning.
\newblock In {\em International Conference on Robotics and Automation (ICRA)},
  pages 6015--6022. IEEE, 2019.

\bibitem{ehsanpour2022jrdb}
Mahsa Ehsanpour, Fatemeh Saleh, Silvio Savarese, Ian Reid, and Hamid
  Rezatofighi.
\newblock Jrdb-act: A large-scale dataset for spatio-temporal action, social
  group and activity detection.
\newblock In {\em Proceedings of the IEEE/CVF Conference on Computer Vision and
  Pattern Recognition (CVPR)}, pages 20983--20992, 2022.

\bibitem{kothari2021human}
Parth Kothari, Sven Kreiss, and Alexandre Alahi.
\newblock Human trajectory forecasting in crowds: A deep learning perspective.
\newblock {\em IEEE Transactions on Intelligent Transportation Systems}, 2021.

\bibitem{le2022pifenet}
Duy~Tho Le, Hengcan Shi, Hamid Rezatofighi, and Jianfei Cai.
\newblock Accurate and real-time 3d pedestrian detection using an efficient
  attentive pillar network.
\newblock {\em IEEE Robotics and Automation Letters}, 8(2):1159--1166, 2022.

\bibitem{luo2018porca}
Yuanfu Luo, Panpan Cai, Aniket Bera, David Hsu, Wee~Sun Lee, and Dinesh
  Manocha.
\newblock Porca: Modeling and planning for autonomous driving among many
  pedestrians.
\newblock {\em IEEE Robotics and Automation Letters}, 3(4):3418--3425, 2018.

\bibitem{martin2021jrdb}
Roberto Martin-Martin, Mihir Patel, Hamid Rezatofighi, Abhijeet Shenoi,
  JunYoung Gwak, Eric Frankel, Amir Sadeghian, and Silvio Savarese.
\newblock Jrdb: A dataset and benchmark of egocentric robot visual perception
  of humans in built environments.
\newblock {\em TPAMI}, 2021.

\bibitem{pang2022simpletrack}
Ziqi Pang, Zhichao Li, and Naiyan Wang.
\newblock Simpletrack: Understanding and rethinking 3d multi-object tracking.
\newblock In {\em European Conference on Computer Vision (ECCV)}, pages
  680--696. Springer, 2022.

\bibitem{rezatofighi2020trustworthy}
Hamid Rezatofighi, Tran Thien~Dat Nguyen, Ba-Ngu Vo, Ba-Tuong Vo, Silvio
  Savarese, and Ian Reid.
\newblock How trustworthy are performance evaluations for basic vision tasks?
\newblock {\em IEEE Transactions on Pattern Analysis and Machine Intelligence
  (TPAMI)}, 2023.

\bibitem{ristani2016idf1}
Ergys Ristani, Francesco Solera, Roger Zou, Rita Cucchiara, and Carlo Tomasi.
\newblock Performance measures and a data set for multi-target, multi-camera
  tracking.
\newblock In {\em European Conference on Computer Vision (ECCV)}, pages 17--35.
  Springer, 2016.

\bibitem{saadatnejad2022sattack}
Saeed Saadatnejad, Mohammadhossein Bahari, Pedram Khorsandi, Mohammad Saneian,
  Seyed-Mohsen Moosavi-Dezfooli, and Alexandre Alahi.
\newblock Are socially-aware trajectory prediction models really
  socially-aware?
\newblock {\em Transportation Research Part C: Emerging Technologies}, 2022.

\bibitem{schuhmacher2008consistent}
Dominic Schuhmacher, Ba-Tuong Vo, and Ba-Ngu Vo.
\newblock A consistent metric for performance evaluation of multi-object
  filters.
\newblock {\em IEEE transactions on signal processing}, 56(8):3447--3457, 2008.

\bibitem{vendrow2023jrdb}
Edward Vendrow, Duy~Tho Le, Jianfei Cai, and Hamid Rezatofighi.
\newblock Jrdb-pose: A large-scale dataset for multi-person pose estimation and
  tracking.
\newblock In {\em Proceedings of the IEEE/CVF Conference on Computer Vision and
  Pattern Recognition (CVPR)}, 2023.

\end{thebibliography}
}

\end{document}